\documentclass[conference]{IEEEtran}
\IEEEoverridecommandlockouts
\usepackage{cite}
\usepackage{bm}
\usepackage{amsmath}               
\usepackage{xpatch}
\usepackage{amsthm}
\usepackage{listings}
\usepackage{verbatim}
\usepackage{amssymb}
\usepackage{mathtools}
\usepackage{dsfont}
\usepackage{hyperref}
\usepackage{xcolor}
\usepackage{float}
\usepackage{lastpage}
\usepackage{fancyhdr}
\usepackage{subfig}
\usepackage[nameinlink,capitalise]{cleveref}
\newtheorem{theorem}{Theorem}
\newtheorem{corollary}{Corollary}
\newtheorem{lemma}{Lemma}
\newtheorem{assumption}{Assumption}

\DeclareMathOperator*{\argmin}{arg\,min}
\hypersetup{
    colorlinks,
    linkcolor={red!0!black},
    citecolor={red!0!black},
    urlcolor={blue!0!black}
}

\newcommand{\multiref}[2]{\ref{#1}-\ref{#2}}
\def\BibTeX{{\rm B\kern-.05em{\sc i\kern-.025em b}\kern-.08em
    T\kern-.1667em\lower.7ex\hbox{E}\kern-.125emX}}
\usepackage[skip=0pt]{caption}
 \newcommand{\Lim}[1]{\raisebox{0.5ex}{\scalebox{0.8}{$\displaystyle \lim_{#1}\;$}}}
\begin{document}
\title{Hierarchical Over-the-Air Federated Edge Learning\\
\thanks{Ozan Ayg{\"u}n's research in this study is supported by Turkcell A.S. within the framework of 5G and Beyond Joint Graduate Support Programme coordinated by Information and Communication Technologies Authority. 

 D. Gunduz acknowledges support from UK EPSRC through CHIST-ERA project CONNECT (CHISTERA-18-SDCDN-001, EPSRC-EP/T023600/1).}
}
\author{\IEEEauthorblockN{Ozan Ayg{\"u}n\textsuperscript{1}, Mohammad Kazemi\textsuperscript{1}, Deniz G{\"u}nd{\"u}z\textsuperscript{2} and Tolga M. Duman\textsuperscript{1}}
\IEEEauthorblockA{\textsuperscript{1}\textit{Dept. of Electrical and Electronics Engineering, Bilkent University}, Ankara, Turkey \\
\textsuperscript{2}\textit{Dept. of Electrical and Electronic Engineering, Imperial College London}, London, UK \\
\{ozan, kazemi, duman\}@ee.bilkent.edu.tr, d.gunduz@imperial.ac.uk}
}
\maketitle
\begin{abstract}
Federated learning (FL) over wireless communication channels, specifically, over-the-air (OTA) model aggregation framework is considered. In OTA wireless setups, the adverse channel effects can be alleviated by increasing the number of receive antennas at the parameter server (PS), which performs model aggregation. However, the performance of OTA FL is limited by the presence of mobile users (MUs) located far away from the PS. In this paper, to mitigate this limitation, we propose hierarchical over-the-air federated learning (HOTAFL), which utilizes intermediary servers (IS) to form clusters near MUs. We provide a convergence analysis for the proposed setup, and demonstrate through theoretical and experimental results that local aggregation in each cluster before global aggregation leads to a better performance and faster convergence than OTA FL.

\end{abstract}


\section{Introduction} \label{sec:Introduction}
Extensive amounts of collected data from various sources such as mobile phones and Internet-of-things (IoT) sensors have enabled the accelerating rise of machine learning (ML) algorithms, aiming to assemble all the data in a cloud server to obtain representative datasets for model training. This, however, brings out growing concerns regarding the privacy, cost, and latency of traditional ML algorithms. Firstly, data owners have become more sensitive about sharing their data; secondly, the increasing quality of data results in higher communication costs; and finally, solutions that work in real-time are faced with latency issues \cite{Lim2020}. To overcome these problems, a decentralized approach called \textit{federated learning} (FL) has been introduced, where the transmission of data is not required since models are trained locally instead of using a centralized server for training \cite{Mcmahan2017}. 

In FL, several data owners called mobile users (MUs) are selected based on some criteria such as their computing capability, data quality, available power, and location \cite{Gunduz2020}. Each MU in the federation trains a local model using its own data and computing power in every iteration. After each MU completes its local stochastic gradient descent (SGD) computation, only the weight updates are sent to a parameter server (PS) that performs model aggregation and sends back the updated global model to MUs for the next iteration. 

Despite its superiority over traditional ML, adverse channel effects in wireless setups and increased communication costs have arisen some concerns about the feasibility of conventional FL in practical scenarios. To address the communication cost concerns, over-the-air (OTA) aggregation \cite{MohammadiAmiri2020} has become a popular method in wireless schemes thanks to its efficient strategy that allocates all the users to the same bandwidth, thereby handling the transmission and aggregation of the gradient updates simultaneously (over the air). For this framework, one approach to deal with the channel effects (particularly when there is no transmit side channel state information) is to increase the number of receive antennas at the PS \cite{Amiri2021a}. Nevertheless, the disparity among the channel gains is still a critical factor when some MUs are far away from the PS. 

Recent developments on FL include device selection algorithms \cite{Amiri2021}, efficient communication schemes \cite{Zhu2020, Zhu2021, MohammadiAmiri2020, Chen2021, tegin, tegin2}, heterogeneity of data \cite{Sery2021}, and power and latency analysis \cite{Dinh2021, Liu2021}. Although \textit{Federated Averaging} \cite{Mcmahan2017} is the most common way to perform global aggregation in error-free setups, OTA communication has been preferred for wireless FL \cite{Amiri2021a, Amiri2021b, Sery2021}. Furthermore, hierarchical federated learning (HFL) has been gaining increasing attention, where the objective is to utilize intermediate servers (IS) to form clusters to reduce communication costs. There exist studies on HFL on latency and power analysis \cite{Abad2020, Liu2020}, resource allocation \cite{Luo2020, Wang2020}, and performance analysis for non-independent and identically distributed (i.i.d.) data \cite{Briggs2020}. However, there is no work on HFL with OTA taking into account practical wireless channel models, which motivates this work.

In order to make distant MUs more resilient to the channel effects, we propose \textit{hierarchical over-the-air federated learning} (HOTAFL), where MUs communicate with their corresponding ISs through wireless links. In this setup, each MU shares its local training result with its corresponding IS through OTA (cluster) aggregation. After several local iterations with the MUs in their clusters, the ISs send the results to the PS to complete the global aggregation, which constitutes one global iteration. We examine the performance of HOTAFL and compare the results with those of the conventional FL and error-free HFL both through analytical results and numerical experiments. The results show that the proposed framework outperforms conventional OTA FL and leads to a better model accuracy and faster convergence.

The paper is organized as follows: in Sections \ref{sec:systemmodel} and \ref{sec:hfl}, we introduce the HOTAFL framework as well as its communication model. In Section \ref{sec:convergenceanalysis}, we provide a convergence analysis of HOTAFL under convexity assumptions on the loss functions. We present our numerical results in Section \ref{sec:simulationresults}, and conclude the paper in Section \ref{sec:conclusions}.
\section{System Model} \label{sec:systemmodel}
\begin{figure}
\begin{center}
\includegraphics[width=.7\columnwidth]{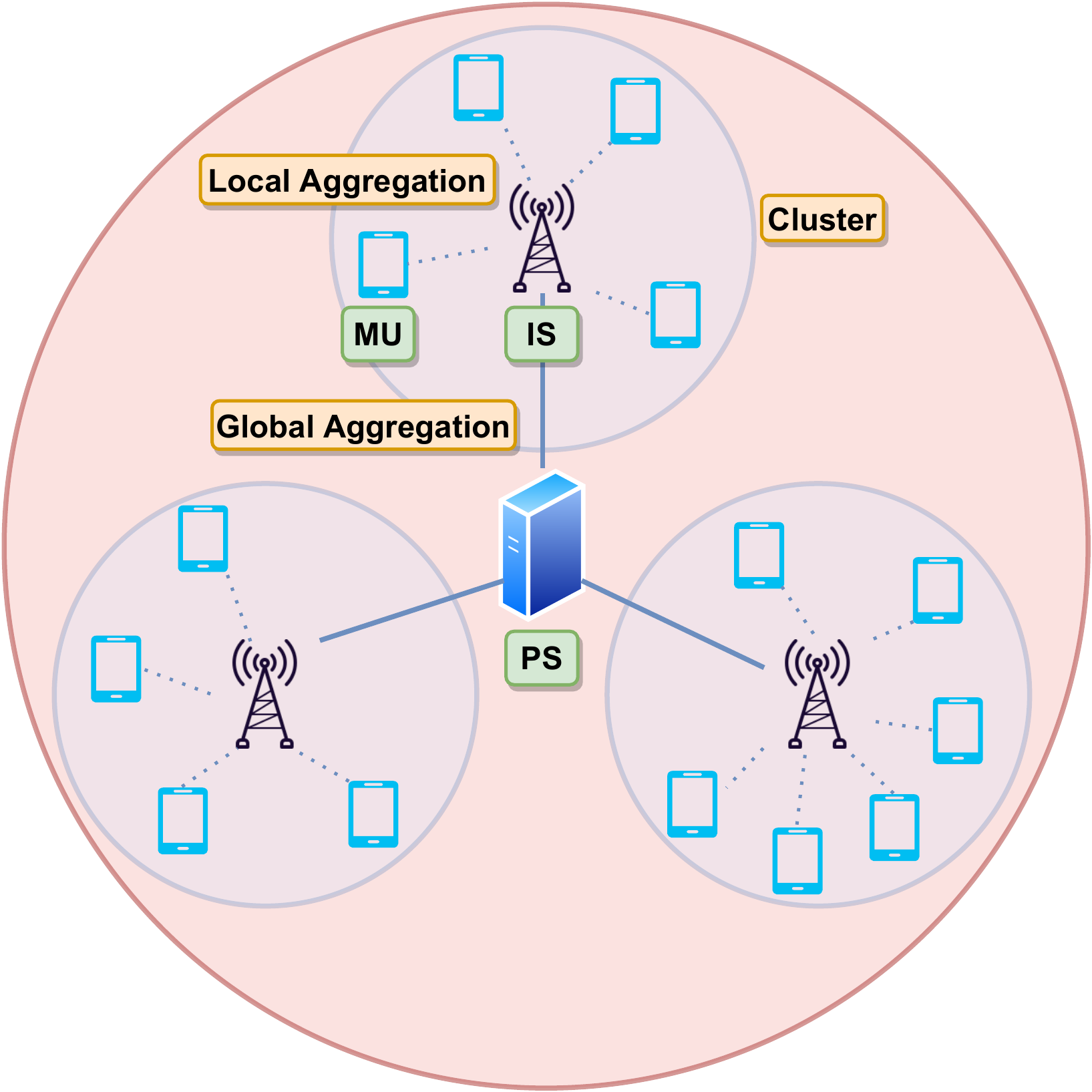} 
\caption{HOTAFL system model.}
\label{fig:system}
\end{center}
\end{figure}

The objective of HOTAFL is to minimize a loss function $F(\bm{\theta})$ with respect to the model weight vector $\bm{\theta} \in \mathbb{R}^{2N}$, where $2N$ is the model dimension. Our system consists of $C$ clusters each containing an IS and $M$ MUs as depicted in Fig. \ref{fig:system}.
The dataset of the $m$-th MU in the $c$-th cluster is denoted as $\mathcal{B}_{m,c}$, and we define $B \triangleq \sum_{c = 1}^{C} \sum_{m = 1}^{M} |\mathcal{B}_{m,c}|$. We have
\begin{equation}
    F(\bm{\theta}) = \sum_{c = 1}^{C} \sum_{m = 1}^{M} \frac{|\mathcal{B}_{m,c}|}{B} F_{m,c}(\bm{\theta}),
    \label{eq:empiricalloss}
\end{equation}
where 
    $F_{m,c}(\bm{\theta}) \triangleq \frac{1}{|\mathcal{B}_{m,c}|} \sum_{u \in \mathcal{B}_{m,c}} f(\bm{\theta},u)$,
with $f(\bm{\theta},u)$ denoting the corresponding loss of $u$-th data sample.

We consider a hierarchical and iterative approach to minimize (\ref{eq:empiricalloss}) consisting of global, local, and user iterations. In every cluster iteration, the MUs carry out $\tau$ user iterations on their own, then send their model updates to their corresponding ISs for local iteration. $I$ local iterations are performed at the IS in every cluster before all the local models are forwarded to the PS for global aggregation. At the $j$-th user iteration of the $i$-th local iteration, the weight update is performed employing stochastic gradient descent (SGD) for the $m$-th user in the $c$-th cluster as follows
\begin{equation}
    \bm{\theta}_{m,c}^{i,j+1}(t) = \bm{\theta}_{m,c}^{i,j}(t) - \eta_{m,c}^{i,j}(t) \nabla F_{m,c} (\bm{\theta}_{m,c}^{i,j}(t), \bm{\xi}_{m,c}^{i,j}(t)),
\end{equation}
where $\eta_{m,c}^{i,j}(t)$ is the learning rate, $\nabla F_{m,c} (\bm{\theta}_{m,c}^{i,j}(t), \bm{\xi}_{m,c}^{i,j}(t))$ denotes the stochastic gradient estimate for the weight vector $\bm{\theta}_{m,c}^{i,j}(t)$ and a randomly sampled batch of data samples $\xi_{m,c}^{i,j}(t)$ from the dataset of the $m$-th user in the $c$-th cluster at the $t$-th global, $i$-th local and $j$-th user iteration. Initially, $\bm{\theta}_{m,c}^{1,1}(t) = \bm{\theta}_{IS,c}^{i}(t), \forall i \in [I]$ where $[I] \triangleq \{ 1,2,\ldots,I \}$, and $\bm{\theta}_{IS,c}^{1}(t) = \bm{\theta}_{PS}(t)$, where $\bm{\theta}_{PS}(t)$ is the global model at the PS at the $t$-th global iteration and $\bm{\theta}_{IS,c}^{i}(t)$ denotes the local model of IS in the $c$-th cluster at the $i$-th local iteration. The purpose of employing ISs is to accumulate the local model differences within each cluster more frequently in smaller areas before obtaining the global model $\bm{\theta}_{PS}(t)$ for the next iteration. Also, note that $\nabla F_{m,c} (\bm{\theta}_{m,c}^{i,j}(t), \bm{\xi}_{m,c}^{i,j}(t))$ is an unbiased estimator of $\nabla F_{m,c} (\bm{\theta}_{m,c}^{i,j}(t))$, i.e., $\mathbb{E}_{\xi} \left[ \nabla F_{m,c} (\bm{\theta}_{m,c}^{i,j}(t), \bm{\xi}_{m,c}^{i,j}(t)) \right] = \nabla F_{m,c} (\bm{\theta}_{m,c}^{i,j}(t)),$
where the expectation is over the randomness due to the SGD.

\section{Hierarchical Over-the-Air FL (HOTAFL)} \label{sec:hfl}
\subsection{Ideal Communication} 
We refer to the case where all the communication among all the units is error-free as the ideal communication scenario. In this case, after performing SGD, each MU calculates its model difference to be sent to its corresponding IS as
\begin{equation}
    \Delta \bm{\theta}_{m,c}^{i}(t) = \bm{\theta}_{m,c}^{i,\tau+1}(t) - \bm{\theta}_{IS,c}^{i}(t).
\end{equation}
Then, the local aggregation at the $c$-th cluster is performed as
\begin{align}
    \Delta \bm{\theta}_{IS,c}^{i}(t) &= \frac{1}{M} \sum_{m = 1}^{M} \Delta \bm{\theta}_{m,c}^{i}(t), \\
    \bm{\theta}_{IS,c}^{i+1}(t) &= \bm{\theta}_{IS,c}^{i}(t) + \Delta \bm{\theta}_{IS,c}^{i}(t).
\end{align}
After completing $I$ local iterations in each cluster, ISs send their model updates to the PS, which can be written as
\begin{equation}
    \Delta \bm{\theta}_{PS,c}(t) = \bm{\theta}_{IS,c}^{I+1}(t) - \bm{\theta}_{PS}(t).
    \label{global1}
\end{equation}
The global update rule is
$\Delta \bm{{\theta}}_{PS}(t) = \frac{1}{C} \sum_{c = 1}^{C} \Delta \bm{\theta}_{PS,c}(t)$.
Using recursion, we can conclude that
\begin{equation}
    \Delta \bm{\theta}_{PS}(t) = \frac{1}{MC} \sum_{c = 1}^{C} \sum_{i = 1}^{I} \sum_{m = 1}^{M} \Delta \bm{\theta}_{m,c}^{i}(t).
    \label{eq: globalmodeldiff}
\end{equation}
After the global aggregation, the model at the PS is updated as
$\bm{\theta}_{PS}(t+1) = \bm{\theta}_{PS}(t) + \Delta \bm{\theta}_{PS}(t)$.
\subsection{OTA Communication}
We now consider the scheme referred as OTA communications, for which the links between the users and the ISs are wireless with OTA aggregation, however, the links between ISs and the PS is assumed to be error-free. Since a common wireless medium is used in local aggregations, noisy versions of the model updates $\Delta \bm{\theta}_{IS,c}(t)$ are received at the ISs. 
In our setup, the ISs are equipped with $K$ antennas, and we assume perfect channel state information (CSI) at the receivers. For the $k$-th antenna, the received signal at the $c$-th IS can be written as\footnote{Note that the setup here can be efficiently implemented in practice using orthogonal frequency-division multiplexing (OFDM).}
\begin{equation}
    \bm{y}_{IS,c,k}^{i}(t) = \sum_{m = 1}^{M} \bm{h}_{m,c,k}^{i}(t) \circ \bm{x}_{m,c,k}^{i}(t) + \bm{z}_{IS,c,k}^{i}(t),
\end{equation}
where $\circ$ denotes the element-wise product, $\bm{x}_{m,c,k}^{i}(t) \in \mathbb{C}^{N}$, $\bm{z}_{IS,c,k}^{i}(t) \in \mathbb{C}^{N}$ with independent and identically distributed (i.i.d.) entries  $ z_{IS,c,k}^{i,n}(t) \sim \mathcal{CN}(0,\sigma_{z}^{2})$. The channel coefficients are modelled as 
    $\bm{h}_{m,c,k}^{i}(t) = \sqrt{\beta_{m,c}} ~ \bm{g}_{m,c,k}^{i}(t)$,
where $\bm{g}_{m,c,k}(t) \in \mathbb{C}^{N}$ with entries $g_{m,c,k}^{i,n}(t) \sim \mathcal{CN}(0,\sigma_{h}^{2})$ (i.e., Rayleigh fading), $\beta_{m,c}$ is the large-scale fading coefficient modeled as $\beta_{m,c} = \left( d_{m,c} \right)^{-p}$, where $p$ represents the path loss exponent, and $d_{m,c}$ denotes the distance between the $m$-th user in the $c$-th cluster and the IS in that cluster. 
\subsubsection{Local Aggregation} 
In OTA communication, in order to increase the spectral efficiency, the model differences are grouped to form a complex vector $\Delta \bm{\theta}_{m,c}^{i,cx} \in \mathbb{C}^{N}$ with the following real and imaginary parts
\begin{subequations}\label{eq:reco}
    \begin{alignat}{2}
        & \!\!\Delta \bm{\theta}_{m,c}^{i,re}(t) \triangleq \left[ \Delta \theta_{m,c}^{i,1}(t), \Delta \theta_{m,c}^{i,2}(t), \ldots, \Delta \theta_{m,c}^{i,N}(t) \right]^{T}, \label{sub-eq-1:reco} \\
        & \!\!\Delta \bm{\theta}_{m,c}^{i,im}(t) \!\triangleq \!\! \left[ \Delta \theta_{m,c}^{i,N+1}\!(t), \Delta \theta_{m,c}^{i,N+2}(t), \ldots, \Delta \theta_{m,c}^{i,2N}\!(t)\! \right]^{\!T}\!\!. \label{sub-eq-2:reco} 
    \end{alignat}
\end{subequations}
Under the assumption that there is no inter-cluster interference, the received signal for the $k$-th antenna in the $c$-th cluster at the $i$-th local iteration can be represented as
\begin{equation}
    \bm{y}_{IS,c,k}^{i}(t) = P_{t} \sum_{m = 1}^{M} \bm{h}_{m,c,k}^{i}(t) \circ \Delta \bm{\theta}_{m,c}^{i,cx}(t) + \bm{z}_{IS,c,k}^{i}(t),
\end{equation}
where $P_{t}$ is the transmit power constant at the $t$-th global iteration. 
Knowing the CSI perfectly, the $c$-th IS combines the received signals as
$\bm{y}_{IS,c}^{i}(t) \!=\! \frac{1}{K} \!\sum_{k = 1}^{K} \! \Big( \!\sum_{m = 1}^{M} \bm{h}_{m,c,k}^{i}(t) \Big)^{\!\!\ast} \! \circ \! \bm{y}_{IS,c,k}^{i}(t)$. For the $n$-th symbol, it can be written as
    \begin{alignat}{2}
        y_{IS,c}^{i,n}(t) \! 
        & = \! \underbrace{P_{t} \! \sum_{m = 1}^{M} \! \Big( \! \frac{1}{K} \! \sum_{k = 1}^{K} \lvert h_{m,c,k}^{i,n}(t) \rvert^{2} \! \Big) \! \Delta \theta_{m,c}^{i,n,cx}(t)}_{\text{$y_{IS,c}^{i,n,sig}(t)$ (signal term)}} \nonumber \\
        & \hspace{0.3cm} +\underbrace{  \frac{P_{t}}{K} \! \sum_{m = 1}^{M} \! \sum_{\substack{m^{\prime} = 1 \\ m^{\prime} \neq m}}^{M} \! \sum_{k = 1}^{K} \! (h_{m,c,k}^{i,n}(t))^{\ast} \! h_{m^{\prime},c,k}^{i,n}(t)  \Delta \theta_{m^{\prime},c}^{i,n,cx}(t)}_{\text{$ y_{IS,c}^{i,n,itf}(t)$ (interference term)}}  \nonumber \\
        & \hspace{0.3cm} +\underbrace{  \frac{1}{K} \! \sum_{m = 1}^{M} \! \sum_{k = 1}^{K}  (h_{m,c,k}^{i,n}(t))^{\ast} z_{c,k}^{i,n}(t)}_{\text{$y_{IS,c}^{i,n,no}(t)$ (noise term)}}.
        \label{eq:sin1}
    \end{alignat}
Aggregated model differences can be recovered by
\begin{subequations}\label{eq:sin11}
    \begin{alignat}{2}
        \Delta \hat{\theta}_{IS,c}^{i,n}(t) &= \frac{1}{P_{t} M \sigma_{h}^{2} \Bar{\beta}_{c}} \operatorname{Re}\{ y_{IS,c}^{i,n}(t) \}, \label{sub-eq-1:sin11} \\
        \Delta \hat{\theta}_{IS,c}^{i,n+N}(t) &= \frac{1}{P_{t} M \sigma_{h}^{2} \Bar{\beta}_{c}} \operatorname{Im}\{ y_{IS,c}^{i,n}(t) \}, \label{sub-eq-2:sin11}
    \end{alignat}
\end{subequations}
where $\Bar{\beta}_{c} = \sum_{m = 1}^{M} \beta_{m,c}$. After estimating the model difference values, the cluster model update is written as
\begin{equation}
    \bm{\theta}_{IS,c}^{i+1}(t) = \bm{\theta}_{IS,c}^{i}(t) + \Delta \bm{\hat{\theta}}_{IS,c}^{i}(t),
\end{equation}
where $\Delta \bm{\hat{\theta}}_{IS,c}^{i}(t) = \big[ \Delta \hat{\theta}_{IS,c}^{i,1}(t) ~ \Delta\hat{\theta}_{IS,c}^{i,2}(t) ~ \cdots ~ \Delta\hat{\theta}_{IS,c}^{i,2N}(t) \big]^{T}$.

\subsubsection{Global Aggregation}
This part is similar to the case of ideal communication. The only difference is that the aggregated signals are estimates of the actual model differences.
Letting $\bm{x}_{PS,c}(t)$ be the transmitted signal from the $c$-th IS, its $n$-th  symbol can be written as
\begin{align}
    x_{PS,c}^{n}(t) = \Delta \theta_{PS,c}^{n}(t) + j \Delta \theta_{PS,c}^{n+N}(t).
    \label{xPS}
\end{align}
Then, using \eqref{global1}, \eqref{eq:sin1}, \eqref{xPS} and recursion, the received signal for $1 \!\leq\! n \!\leq\! N$ (similarly for $N\!+\!1 \!\leq\! n \!\leq\! 2N$) can be written as

\begin{align}
    &y_{PS}^{n}(t) = \sum_{c = 1}^{C} x_{PS,c}^{n}(t) \\
    & = \!\!\! \underbrace{\sum_{c = 1}^{C} \! \sum_{i = 1}^{I} \! \frac{\operatorname{Re} \big\{ y_{IS,c}^{i,n,sig}(t)\big\}}{P_{t}M\sigma_{h}^{2}}}_{y_{PS,1}^{n}(t)} \!+\! \underbrace{\sum_{c = 1}^{C} \! \sum_{i = 1}^{I} \! \frac{\operatorname{Re} \big\{y_{IS,c}^{i,n,itf}(t)\big\}}{P_{t}M\sigma_{h}^{2}}}_{y_{PS,2}^{n}(t)} \nonumber \\
    &\quad + \underbrace{\sum_{c = 1}^{C} \! \sum_{i = 1}^{I} \! \frac{\operatorname{Re} \big\{y_{IS,c}^{i,n,no}(t)\big\}}{P_{t}M\sigma_{h}^{2}}}_{y_{PS,3}^{n}(t)}\!. 
\end{align}
The received signal at the PS is then recovered as
$\Delta \hat{\theta}_{PS}^{n}(t) = \frac{1}{C} \operatorname{Re}\{ y_{PS}^{n}(t) \}$, $\Delta \hat{\theta}_{PS}^{n+N}(t) = \frac{1}{C} \operatorname{Im}\{ y_{PS}^{n}(t) \}$.
Finally, the global aggregation is performed as
\begin{align} 
    \bm{\theta}_{PS}(t+1) = \bm{\theta}_{PS}(t) + \Delta \bm{\hat{\theta}}_{PS}(t), \label{eq:globalaggregation}
\end{align}
where $\Delta \bm{\hat{\theta}}_{PS}(t) = \big[ \Delta \hat{\theta}_{PS}^{1}(t) ~ \Delta\hat{\theta}_{PS}^{2}(t) ~ \cdots ~ \Delta\hat{\theta}_{PS}^{2N}(t) \big]^{T}$.
\section{Convergence Analysis} \label{sec:convergenceanalysis}
In this section, we present a convergence analysis of the proposed HOTAFL algorithm. Define the optimal solution as $\bm{\theta}^{\ast} \triangleq \argmin_{\bm{\theta}} F(\bm{\theta})$,
 the minimum values of the total and the local loss functions as $F^{*} = F(\bm{\theta}^{*})$ and $F_{m,c}^{*}$, respectively, and the bias in the dataset as
    $\Gamma \triangleq F^{*} - \sum_{c = 1}^{C} \sum_{m = 1}^{M} \frac{B_{m,c}}{B} F_{m,c}^{*} \geq 0$.
In addition, assume that the learning rate of the overall system does not change in local iterations, i.e., $\eta_{m,c}^{i,j}(t) = \eta(t)$. Therefore, we can write the global update rule as
\begin{align}
    \bm{\theta}_{m,c}^{i,j+1}(t) & = \bm{\theta}_{m,c}^{i,j}(t) - \eta(t) \nabla F_{m,c} (\bm{\theta}_{m,c}^{i,j}(t), \bm{\xi}_{m,c}^{i,j}(t)),
\end{align}
which can also be written as
\begin{equation}
    \bm{\theta}_{m,c}^{i,j+1}(t) - \bm{\theta}_{m,c}^{i,1}(t) = - \eta(t) \sum_{l = 1}^{j} \nabla F_{m,c} (\bm{\theta}_{m,c}^{i,l}, \bm{\xi}_{m,c}^{i,l}(t)).
    \label{eq: useriteration}
\end{equation}
\begin{assumption} \label{assumption1}
All the loss functions are L-smooth and $\mu$-strongly convex; i.e., $\forall \bm{v},\bm{w} \in \mathbb{R}^{2N}$, $\forall m \in [M], \forall c \in [C]$,
\begin{align}
    \!\!F_{m,c}(\bm{v}) \!-\! F_{m,c}(\bm{w}) & \!\leq\! \langle \bm{v} \!-\! \bm{w}, \! \nabla\! F_{m,c} (\bm{w}) \rangle \!+\! \frac{L}{2} \!\left\| \bm{v} - \bm{w} \right\|_{2}^{2}\!, \\
    \!\!F_{m,c}(\bm{v}) \!-\! F_{m,c}(\bm{w}) & \!\geq\! \langle \bm{v} \!-\! \bm{w},  \!\nabla\! F_{m,c} (\bm{w}) \rangle \!+\! \frac{\mu}{2} \!\left\| \bm{v} - \bm{w} \right\|_{2}^{2}\!.
\end{align}
\end{assumption}
\begin{assumption} \label{assumption2}
The expected value of the squared $l_{2}$ norm of the stochastic gradients are bounded; i.e., $\forall j \in [\tau], i \in [I]$, 
\begin{equation}
    \mathbb{E}_{\xi} \Big[  \left\| \nabla F_{m,c} (\bm{\theta}_{m,c}^{i,j}(t), \bm{\xi}_{m,c}^{i,j}(t)) \right\|_{2}^{2} \Big] \leq G^{2},
\end{equation}
which translates to $\forall n \!\in\! [2N]$,
    $\mathbb{E}_{\xi} \!\!\left[  \nabla \!F_{m,c} (\theta_{m,c}^{i,j,n}, \xi_{m,c}^{i,j,n}(t)) \!\right] \!\!\leq\! G$.
\end{assumption}
\begin{theorem} \label{theorem1}
In HOTAFL, for $0 \leq \eta(t) \leq min\{ 1, \frac{1}{\mu \tau I} \}$, the global loss function can be upper bounded as
\begin{align}
    & \mathbb{E} \big[ \left\| \bm{\theta}_{PS}(t) - \bm{\theta}^{*} \right\|_{2}^{2} \big] \nonumber \\ 
    & \leq \! \bigg( \prod_{a = 1}^{t-1} \! X(a) \! \bigg) \! \left\| \bm{\theta}_{PS}(0) \! - \! \bm{\theta}^{*} \right\|_{2}^{2} \! + \! \sum_{b = 1}^{t-1} \! Y(b) \! \prod_{a = b+1}^{t-1} \! X(a), \label{ourt1}
\end{align}
where $X(a) = \left( 1 - \mu \eta(a) I \left( \tau - \eta(a) (\tau - 1) \right) \right)$ and

\begin{align} 
     Y(a) \!=& \frac{\eta^2(a) \tau^2 G^2 I}{M^2 C^2} \!\!\sum_{m_{1} = 1}^{M} \sum_{c_{1} = 1}^{C} \!\Big( \frac{\beta_{m_{1},c_{1}}^{2}}{K \Bar{\beta_{c_{1}}^{2}}}  \!+\!\!  \Big( \! \sum_{m_{2} = 1}^{M} \sum_{c_{2} = 1}^{C}  A_{1} I \!\Big) \!\! \Big) \nonumber \\
    & + \sum_{m = 1}^{M} \sum_{\substack{m^{\prime} = 1 \\ m^{\prime} \neq m}}^{M} \sum_{c = 1}^{C}  \frac{\eta^2(a) \tau^2 G^2 I \beta_{m,c} \beta_{m^{\prime},c}}{M^2 C^2 K \Bar{\beta}_{c}^{2}} \nonumber \\
    & + \frac{\sigma_{z}^{2} I N}{P_{a}^{2} M^2 C^2 K \sigma_{h}^{2}} \sum_{m = 1}^{M} \sum_{c = 1}^{C} \frac{\beta_{m,c}}{\Bar{\beta}_{c}^{2}} \nonumber \\
    & + \left( 1 + \mu (1- \eta(a) \right) \eta^{2}(a) I G^{2} \frac{\tau (\tau - 1) (2 \tau - 1)}{6} \nonumber \\
    & + \eta^{2}(a) I (\tau^{2} + \tau - 1) G^{2} + 2 \eta(a) I (\tau - 1) \Gamma, \label{eq:xtyt}
\end{align}
with $A_{1}  = 1 - \frac{\beta_{m_{1},c_{1}}}{\Bar{\beta}_{c_{1}}} - \frac{\beta_{m_{2},c_{2}}}{\Bar{\beta}_{c_{2}}} + \frac{ \beta_{m_{1},c_{1}} \beta_{m_{2},c_{2}}}{ \Bar{\beta}_{c_{1}} \Bar{\beta}_{c_{2}}}$.
\end{theorem}
\begin{IEEEproof} 
Let us define auxiliary variable
   $ \bm{v}(t+1) = \bm{\theta}_{PS}(t) + \Delta \bm{\theta}_{PS}(t)$.
Then, we have
\begin{align}
    &\!\!\left\| \bm{\theta}_{PS}(t\!+\!1) \!-\! \bm{\theta}^{\ast} \right\|_{2}^{2} \nonumber 
    \!=\! \left\| \bm{\theta}_{PS}(t\!+\!1) \!-\! \bm{v}(t\!+\!1) + \bm{v}(t\!+\!1) \!-\! \bm{\theta}^{\ast} \right\|_{2}^{2} \nonumber \\
    &\!\!\quad= \left\| \bm{\theta}_{PS}(t+1) - \bm{v}(t+1) \right\|_{2}^{2} + \left\| \bm{v}(t+1) - \bm{\theta}^{\ast} \right\|_{2}^{2} \nonumber \\
    &\!\!\quad\quad+ 2 \langle \bm{\theta}_{PS}(t+1) - \bm{v}(t+1) , \bm{v}(t+1) - \bm{\theta}^{\ast}  \rangle. \label{eq: theorem1argument}
\end{align}
Next, we provide upper bounds on the three terms of (\ref{eq: theorem1argument}).
\begin{lemma} \label{convproof1} 
We have
\begin{align}
    & \mathbb{E} \Big[ \big\| \bm{\theta}_{PS}(t+1) - \bm{v}(t+1) \big\|_{2}^{2} \Big] \nonumber \\
    & \leq \frac{\eta^2(t) \tau^2 G^2 I}{M^2 C^2} \sum_{m_{1} = 1}^{M} \sum_{c_{1} = 1}^{C} \Big( \frac{\beta_{m_{1},c_{1}}^{2}}{K \Bar{\beta_{c_{1}}^{2}}}  +  \Big( \sum_{m_{2} = 1}^{M} \sum_{c_{2} = 1}^{C}  A_{1} I \Big)  \Big) \nonumber \\
    &\quad + \sum_{m = 1}^{M} \sum_{\substack{m^{\prime} = 1 \\ m^{\prime} \neq m}}^{M} \sum_{c = 1}^{C}  \frac{\eta^2(t) \tau^2 G^2 I \beta_{m,c} \beta_{m^{\prime},c}}{M^2 C^2 K \Bar{\beta}_{c}^{2}} \nonumber \\
    & \quad + \frac{\sigma_{z}^{2} I N}{P_{t}^{2} M^2 C^2 K \sigma_{h}^{2}} \sum_{m = 1}^{M} \sum_{c = 1}^{C} \frac{\beta_{m,c}}{\Bar{\beta}_{c}^{2}}.
\end{align}
\end{lemma}
\begin{IEEEproof} 
See Appendix \ref{sec:appendix1}.
\end{IEEEproof}
\begin{lemma} \label{convproof2}
We have
\begin{align}
    & \!\mathbb{E}\! \Big[ \!\big\| v(t\!\!+\!\!1) \!-\! \bm{\theta}^{*} \!\big\|_{2}^{2} \!\Big] 
    \!\!\leq\! \!\left( 1 \!-\! \mu \eta(t) \!I \!\left( \tau \!-\! \eta(t) (\!\tau \!\!-\!\! 1) \!\right) \!\right) \!\mathbb{E} \!\Big[ \!\big\| \bm{\theta}_{\!PS}(t) \!-\! \bm{\theta}^{*}  \!\big\|_{2}^{2} \!\Big] \nonumber \\
    & \quad + \left( 1 + \mu (1- \eta(t) \right) \eta^{2}(t) I G^{2} \frac{\tau (\tau - 1) (2 \tau - 1)}{6} \nonumber \\
    & \quad + \eta^{2}(t) I (\tau^{2} + \tau - 1) G^{2} + 2 \eta(t) I (\tau - 1) \Gamma.
\end{align}
\end{lemma}
\begin{IEEEproof}
The proof is similar to that of Lemma 2 in \cite{Amiri2021a}.
\end{IEEEproof}
\begin{lemma} \label{convproof3}
    $\mathbb{E} \left[ \langle \bm{\theta}_{PS}(t+1) - \bm{v}(t+1) , \bm{v}(t+1) - \bm{\theta}^{\ast}  \rangle \right] = 0$.
\end{lemma}
\begin{IEEEproof}
We have
    $ \mathbb{E} \!\left[ \langle \bm{\theta}_{\!PS}(t\!+\!1) \!-\! \bm{v}(t\!+\!1) , \bm{v}(t\!+\!1) \!-\! \bm{\theta}^{\ast}  \rangle \right] \!= \! \mathbb{E} \! \left[ \! \langle \Delta \bm{\hat{\theta}}_{\!PS}(t) \! - \! \Delta \bm{\theta}_{\!PS}(t), \! \bm{\theta}_{\!PS}(t) \! + \! \Delta \bm{\theta}_{\!PS}(t) \! - \! \bm{\theta}^{*} \rangle \! \right]$.
Then, knowing that channel realizations are independent of the user and cluster updates at the same global iteration $t$, we have
   $\mathbb{E} \! \left[ \! \langle \Delta \bm{\hat{\theta}}_{\!PS}(t) \! - \! \Delta \bm{\theta}_{\!PS}(t), \! \bm{\theta}_{\!PS}(t) \! + \! \Delta \bm{\theta}_{\!PS}(t) \! - \! \bm{\theta}^{*} \rangle \! \right] \! = \! 0$.
\end{IEEEproof}
Recursively iterating through the results of Lemmas \ref{convproof1}, \ref{convproof2}, and \ref{convproof3} concludes the theorem.
\end{IEEEproof}
\begin{corollary}
Assuming $L$-smoothness, after $T$ global iterations, the loss function can be upper bounded as
\begin{align} \label{eq:cor1}
    & \!\!\mathbb{E} \left[ F(\bm{\theta}_{PS}(T)) - F^{*} \right] \leq \frac{L}{2} \mathbb{E} \left[ \left\| \bm{\theta}_{PS}(T) - \bm{\theta}^{*} \right\|_{2}^{2} \right] \nonumber \\
    & \!\!\leq \! \frac{L}{2} \! \bigg( \prod_{n = 1}^{T-1} \!\! X(n) \!\! \bigg) \!\! \left\| \bm{\theta}_{PS}(0) \! - \! \bm{\theta}^{*} \right\|_{2}^{2} \! + \! \frac{L}{2} \! \sum_{p = 1}^{T-1} \! Y(p) \!\!\! \!\prod_{n = p+1}^{T-1} \!\!\!\! X(n).
\end{align}
\end{corollary}
\noindent {\textbf{Remark.}} Since the third term in $Y(a)$ is independent of $\eta(a)$, even for $\Lim{t \to \infty} \eta(t) = 0$, we have $\Lim{t \to \infty} \mathbb{E} [F(\bm{\theta}_{PS}(t))] - F^{*} \neq 0$. This term is also inversely proportional to $M$, $C$, and $K$.
\section{Simulation Results} \label{sec:simulationresults}
We consider a hierarchical system with one PS and $C = 4$ non-overlapping clusters, each containing one IS with $K = 5MC$ receive antennas and $M = 5$ MUs. Users are randomly placed in the clusters in such a way that their distance to the PS is between 0.5 and 3, while having a distance between 0.5 and 1 with their corresponding IS. We also define
    $\alpha = \frac{\sum_{m = 1}^{M} \sum_{c = 1}^{C} d_{m,c}}{\sum_{d = 1}^{D} d_{d}}$
as a measure of relative closeness of the MUs to their corresponding IS compared to the PS, where $d_{m,c}$ is the distance between $m$-th user in $c$-th cluster to the $c$-th IS and $d_{d}$ is the distance between the $d$-th user and the PS. $\alpha$ is set to 0.4 in the simulations.
We use MNIST \cite{mnist} and CIFAR-10 \cite{cifar10} datasets with Adam optimizer \cite{adam}, and considered both i.i.d. and non-i.i.d. data distributions. In the i.i.d. case, data samples are randomly distributed among MUs, while in the non-i.i.d. case, the training data is divided into $5MC$ groups each consisting of data with only one label. Then, 5 groups are assigned to each user randomly. For CIFAR10, we use the neural network given in \cite{Amiri2021a} with $2N=307498$ whereas for MNIST, we employ a one-layer neural network with 784 input and 10 output neurons with $2N=7850$. 

Three scenarios are considered: baseline with error-free transmissions, FL with OTA aggregation over a wireless medium, and HOTAFL. We set the total number of global iterations $T$ to 200, the mini-batch size to $|\bm{\xi}_{m,c}^{i}(t)| = 500$, $\sigma_{h}^{2} = 1$ and the path loss exponent $p$ to 4. The noise variance is $\sigma_{z}^{2} = 10$ for the MNIST, $\sigma_{z}^{2} = 1$ for the CIFAR-10 training. Also, the power multiplier is set to $P_{t} = 1 + 10^{-2} t$ for HOTAFL, $P_{t} = 1.5 + 10^{-2} t$ for conventional FL, $t \in [T]$. 

\begin{figure}[t]
  \centering
  \includegraphics[width=.8\columnwidth]{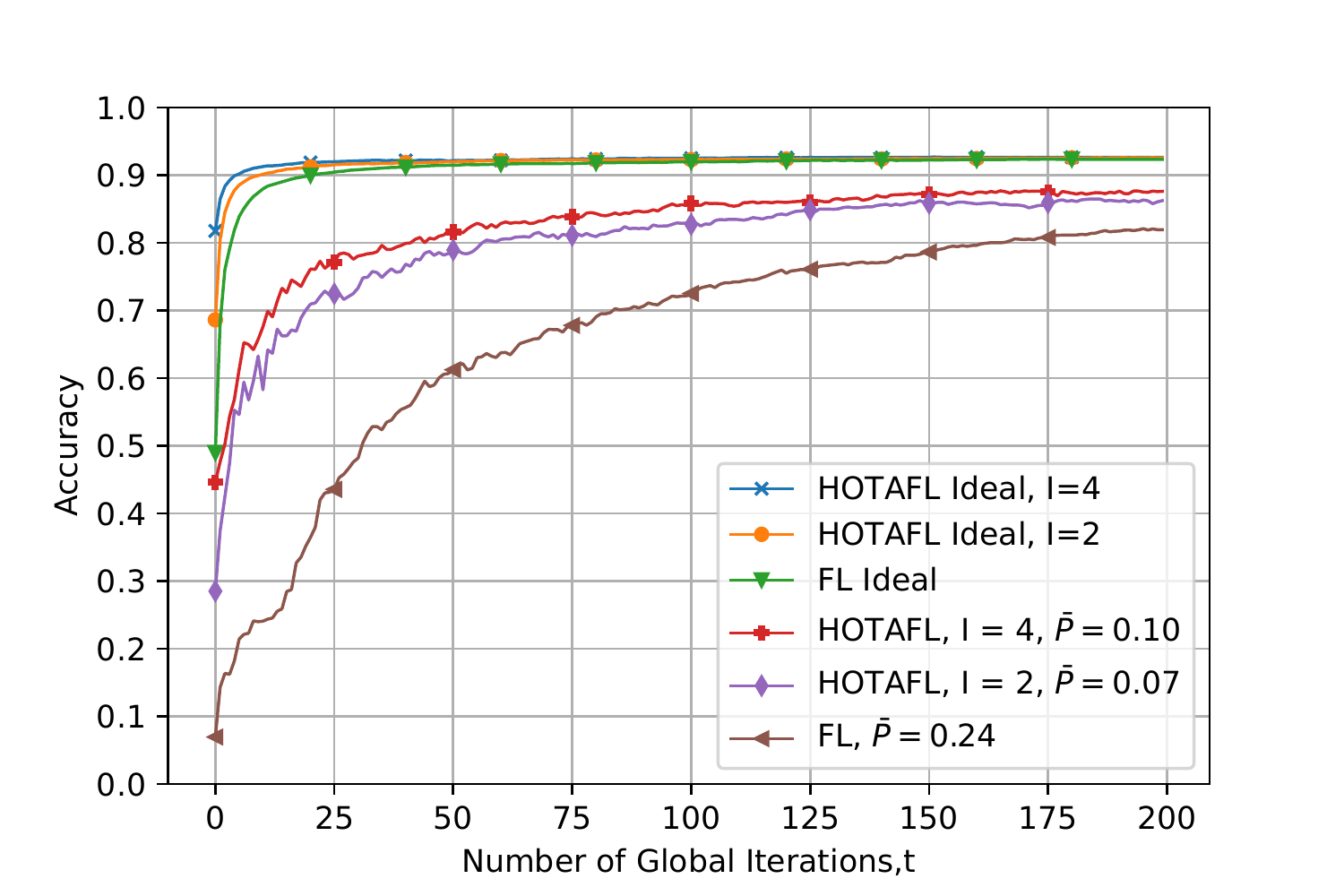} 
\caption{Test accuracy for i.i.d. MNIST data with $\tau = 1$.}
\label{fig:mnistiid}
\vspace{2mm}
  \centering
 \includegraphics[width=.8\columnwidth]{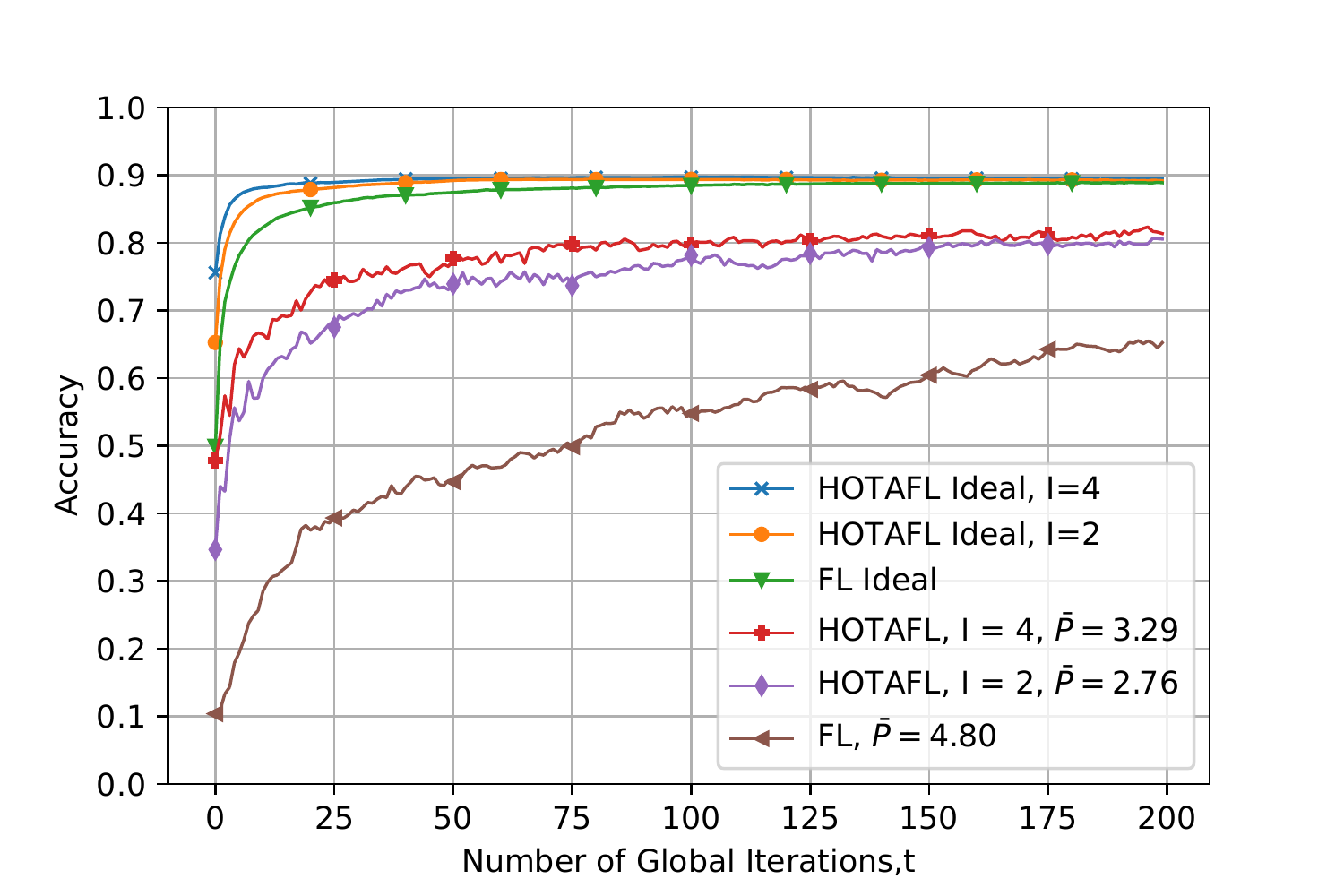} 
\caption{Test accuracy for non-i.i.d. MNIST data with $\tau = 3$.}
\label{fig:mnistnoniid}
\end{figure}

\begin{figure}[t]
  \centering
  \includegraphics[width=.8\columnwidth]{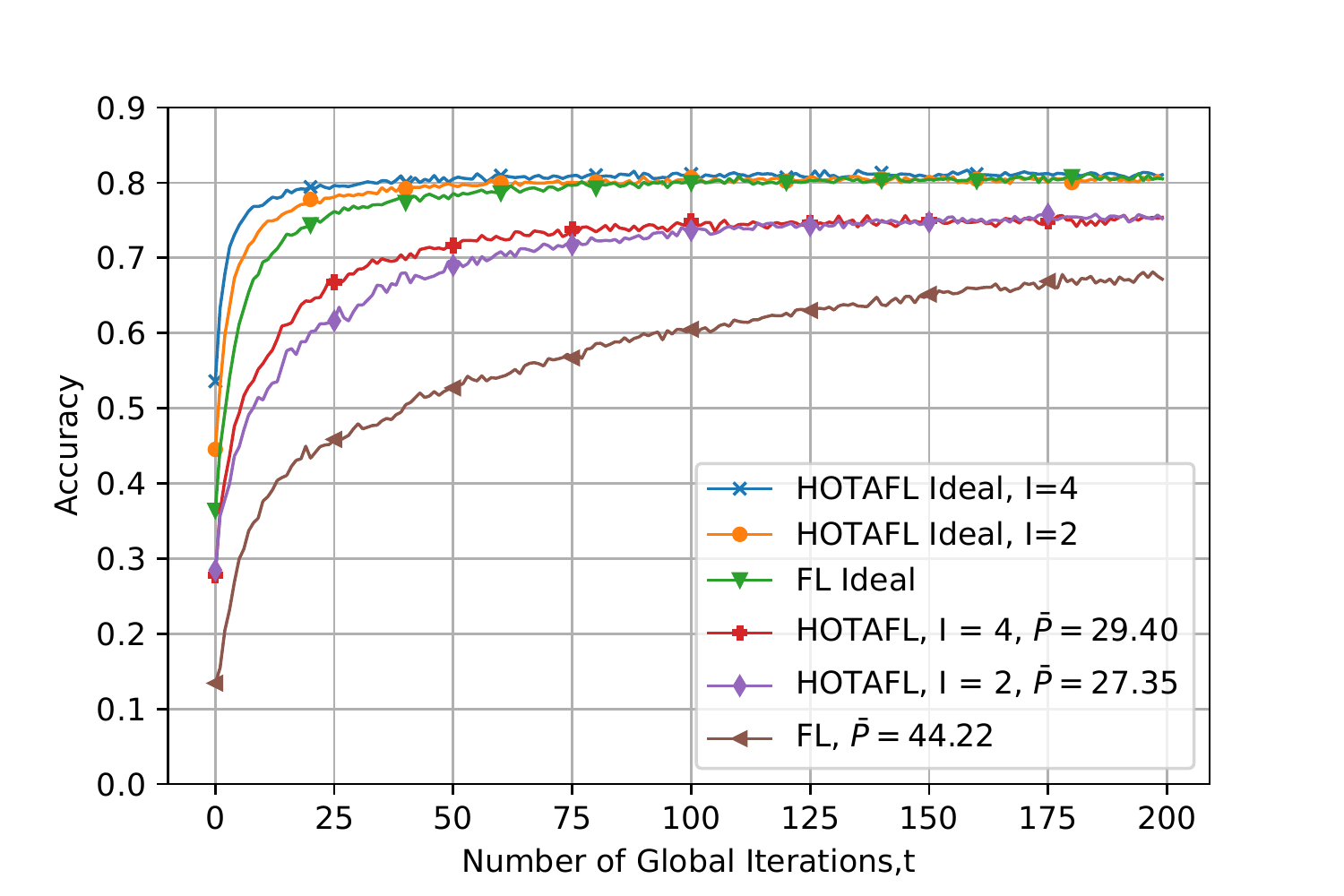} 
\caption{Test accuracy for i.i.d. CIFAR-10 data with $\tau = 5$.}
\label{fig:cifar}
\vspace{2mm}
  \centering
  \includegraphics[width=.8\columnwidth]{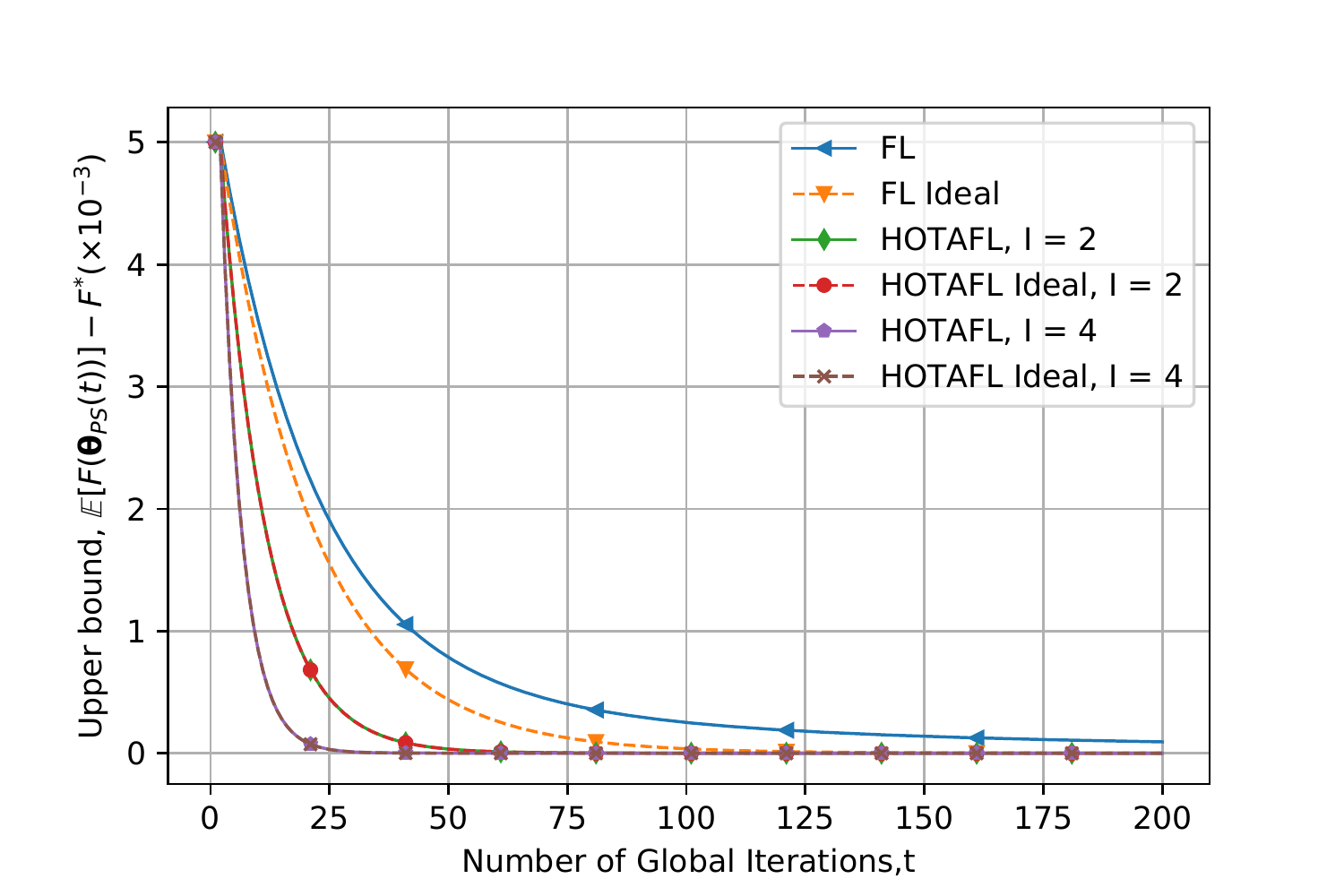} 
\caption{Convergence rate for i.i.d. MNIST data with $\tau =1$. }
\label{fig:upperbound}
\end{figure}

Accuracy plots are presented in Figs. \multiref{fig:mnistiid}{fig:cifar}, where $\Bar{P}$ is the average transmit power. The results show that bringing the servers closer to the users enhances learning accuracy significantly. One reason for the improved performance is that the cluster structure enables the MUs share their model differences with a local server closer than the PS, reducing the adverse effects of the large-scale wireless channel effects. Another reason is that MUs receive updated models even without communicating with the PS due to local aggregations. We also observe that although more initial power is given to FL, the user updates do not reflect on the global model as much as HOTAFL does due to the effects of the wireless channel.  More local iterations enables faster convergence but uses more transmit power due to the accumulating nature of IS. Even though the noise variance is high when compared to the transmit power, deploying $5MC = 100$ receive antennas almost mitigates the noise and the interference terms \cite{Amiri2021a}. Increasing $\tau$ compensates the accuracy under more complex data structure.
In Fig. \ref{fig:upperbound}, we compare the convergence rates of conventional FL and HOTAFL using the upper bound in (\ref{eq:cor1}), with $2N = 7850, L = 10, \mu = 1, G^{2} = 1, \Gamma = 1, \eta(t) = 5 \! \cdot \! 10^{-2} \! - \! 2 \cdot  10^{-5}t, P_{t} = 1 + 10^{-2}t, \beta = 3, \left\| \bm{\theta}_{PS}(0) - \bm{\theta}^{*} \right\|_{2}^{2} = 10^{3}$. It can be seen that the convergence rate of HOTAFL is very close to that of the ideal case, and it becomes almost the same when the number of local iterations is increased. 

\section{Conclusions} \label{sec:conclusions}
In this work, we have proposed HOTAFL where ISs are employed to create clusters to bring the server-side closer to the areas where MUs are more densely located. Our framework includes OTA cluster aggregations, where the MUs send their model updates to the ISs through a wireless channel with path loss and fading. We have examined the performance and convergence rate of HOTAFL through theoretical limits as well as model training where MNIST and CIFAR-10 datasets are used with both i.i.d. and non-i.i.d. data distributions. The results show that employing a cluster-based hierarchical model outperforms the conventional FL.

\appendices
\section{} 
\label{sec:appendix1}
For the $n$-th symbol, we have
    $\Delta \hat{\theta}_{PS}^{n}(t) = \sum_{l = 1}^{3} \Delta \hat{\theta}_{PS,l}^{n}(t)$,
so using the independence of channel coefficients, we have
\begin{align}
    &\!\mathbb{E} \big[ || \bm{\theta}_{PS}(t\!+\!1) \!-\! \bm{v}(t\!+\!1) ||_{2}^{2} \big] 
      = \mathbb{E} \big[ \big\| \Delta \hat{\bm{\theta}}_{PS}(t)  \!-\! \Delta \bm{\theta}_{PS}(t) \big\|_{2}^{2} \big] \nonumber\\
    & \!\!= \!\!\sum_{n = 1}^{2N} (\mathbb{E} \big[ \!\big(\! \Delta \hat{\theta}_{PS,1}^{n}(t)  \!-\! \Delta \theta_{PS}^{n}(t) \big)^{\!2} \big] \!\!+\!\! \sum_{l = 2}^{3} \mathbb{E} \big[ \!\big( \!\Delta \hat{\theta}_{PS,l}^{n}(t) \big)^{\!2} \big]\!.
\end{align}
In the following lemmas, we will bound each of these terms.
\begin{lemma} \label{convproof11} 
We have
\begin{align}
    & \sum_{n = 1}^{2N} \mathbb{E} \big[ \big( \Delta \hat{\theta}_{PS,1}^{n}(t)  - \Delta \theta_{PS}^{n}(t) \big)^{2} \big] \nonumber \\
    & = \frac{1}{M^2 C^2} \sum_{m_{1} = 1}^{M} \sum_{c_{1} = 1}^{C} \sum_{i_{1} = 1}^{I} \Big( \frac{\beta_{m_{1},c_{1}}^{2}}{K \Bar{\beta_{c_{1}}^{2}}} \mathbb{E} \big[ \big\| \Delta \bm{\theta}_{m_{1},c_{1}}^{i_{1}}(t) \big\|_{2}^{2} \big] \big. \nonumber \\
    & \big. + \! \Big( \sum_{m_{2} = 1}^{M} \! \sum_{c_{2} = 1}^{C} \! \sum_{i_{2} = 1}^{I} \! \sum_{n = 1}^{2N} \! A_{1} \mathbb{E} \big[ \Delta \theta_{m_{1},c_{1}}^{i_{1},n}(t) \Delta \theta_{m_{2},c_{2}}^{i_{2},n}(t) \big] \! \Big) \! \Big),
    \label{lemma4}
\end{align}
where 
   $A_{1} = 1 - \frac{\beta_{m_{1},c_{1}}}{\Bar{\beta}_{c_{1}}} - \frac{\beta_{m_{2},c_{2}}}{\Bar{\beta}_{c_{2}}} + \frac{ \beta_{m_{1},c_{1}} \beta_{m_{2},c_{2}}}{ \Bar{\beta}_{c_{1}} \Bar{\beta}_{c_{2}}}$
\end{lemma}
\begin{IEEEproof} 
Using (\ref{eq: globalmodeldiff}) and (\ref{eq:sin11}), we have
\begin{align}
    &\mathbb{E} \big[ \big( \Delta \hat{\theta}_{PS,1}^{n}(t)  - \Delta \theta_{PS}^{n}(t) \big)^{2} \big] \nonumber \\
    & = \mathbb{E} \Big[ \frac{1}{M^2 C^2} \sum_{m_{1} = 1}^{M} \sum_{m_{2} = 1}^{M} \sum_{c_{1} = 1}^{C} \sum_{c_{2} = 1}^{C} \sum_{i_{1} = 1}^{I} \sum_{i_{2} = 1}^{I} \Delta \theta_{m_{1},c_{1}}^{i_{1},n}(t)  \Big. \nonumber \\ 
    & \quad \Big. \times \Delta \theta_{m_{2},c_{2}}^{i_{2},n}(t)  \Big( 1 - \frac{1}{K \sigma_{h}^{2} \Bar{\beta}_{c_{1}}} \sum_{k_{1} = 1}^{K} |h_{m_{1},c_{1},k_{1}}^{i_{1},n}(t)|^{2} \Big. \Big. \nonumber \\
    & \quad \Big. \Big. - \frac{1}{K \sigma_{h}^{2} \Bar{\beta}_{c_{2}}} \sum_{k_{2} = 1}^{K} |h_{m_{2},c_{2},k_{2}}^{i_{2},n}(t)|^{2} \Big. \Big. \nonumber \\
    & \quad \Big. \Big. + \! \frac{1}{K^{2} \sigma_{h}^{4} \Bar{\beta}_{c_{1}}^{2}} \!\! \sum_{k_{1} = 1}^{K} \! \sum_{k_{2} = 1}^{K} \!\! |h_{m_{1},c_{1},k_{1}}^{i_{1},n}\!(t)|^{2}  |h_{m_{2},c_{2},k_{2}}^{i_{2},n}\!(t)|^{2} \! \Big) \! \Big]\!. \label{eq:lemma4expanded}
\end{align}
Summing over all the symbols and using the independence of channel coefficients result in \eqref{lemma4}.
\end{IEEEproof}
\begin{lemma} \label{convproof12} 
We have
\begin{align}
    &\sum_{n = 1}^{2N} \mathbb{E} \big[ \big( \Delta \hat{\theta}_{PS,2}^{n}(t)  \big)^{2} \big] \nonumber \\
     &= \sum_{m = 1}^{M} \sum_{\substack{m^{\prime} = 1 \\ m^{\prime} \neq m}}^{M} \!\sum_{c = 1}^{C} \sum_{i = 1}^{I} \frac{\beta_{m,c} \beta_{m^{\prime},c}}{M^2 C^2 K \Bar{\beta}_{c}^{2}} \mathbb{E} \big[ \left\| \Delta \bm{\theta}_{m^{\prime},c}^{i}(t) \right\|_{2}^{2} \big].
     \label{lemma5}
\end{align}
\end{lemma}
\begin{IEEEproof}
For $1 \leq n \leq N$, using the independence of channel coefficients, we have
\begin{align}
    &\mathbb{E} \big[ \big( \Delta \hat{\theta}_{PS,2}^{n}(t) \big)^{2} \big] = \mathbb{E} \Big[ \Big( \sum_{m = 1}^{M} \sum_{\substack{m^{\prime} = 1 \\ m^{\prime} \neq m}}^{M} \sum_{c = 1}^{C} \sum_{i = 1}^{I} \frac{1}{MCK \sigma_{h}^{2} \Bar{\beta}_{c}} \Big. \Big. \nonumber \\
    & \hspace{0.5cm} \Big. \Big. \times \sum_{k = 1}^{K} \operatorname{Re} \big\{ \big(  h_{m,c,k}^{i,n}(t) \big)^{*} h_{m^{\prime},c,k}^{i,n}(t) \Delta \theta_{m^{\prime},c}^{i,n}(t) \big\} \Big)^{2} \Big] \nonumber\\
    & = \mathbb{E} \Big[ \sum_{m = 1}^{M} \sum_{\substack{m^{\prime} = 1 \\ m^{\prime} \neq m}}^{M} \sum_{c = 1}^{C} \sum_{i = 1}^{I} \frac{\beta_{m,c} \beta_{m^{\prime},c}}{2 M^2 C^2 K \Bar{\beta}_{c}^{2}} \Big. \nonumber \\
    & \enskip \Big. \times \big( \big( \Delta \theta_{m^{\prime},c}^{i,n}(t) \big)^{2} + \big( \Delta \theta_{m^{\prime},c}^{i,n+N}(t) \big)^{2} +  \Delta \theta_{m,c}^{i,n}(t) \Delta \theta_{m^{\prime},c}^{i,n}(t) \big. \Big. \nonumber \\
    & \hspace{3cm} \Big. \left. - \Delta \theta_{m,c}^{i,n+N}(t) \Delta \theta_{m^{\prime},c}^{i,n+N}(t) \right) \Big]
\end{align}
Obtaining the expressions for $N+1 \leq n \leq 2N$ in a similar manner and 
combining the two, results in \eqref{lemma5}.
\end{IEEEproof}
\begin{lemma} \label{convproof13} 
\begin{align}
    \sum_{n = 1}^{2N} \! \mathbb{E}  \big[  \big(  \Delta \hat{\theta}_{PS,3}^{n}(t)  \big)^{2}  \big] \! = \! \frac{\sigma_{z}^{2} I N}{P_{t}^{2} M^2 C^2 K \sigma_{h}^{2}} \! \sum_{m = 1}^{M} \! \sum_{c = 1}^{C} \! \frac{\beta_{m,c}}{\Bar{\beta}_{c}^{2}}.
\end{align}
\end{lemma}
\begin{IEEEproof}
Using the independence of channel coefficients, for $1 \leq n \leq N$, we have
\begin{align}
    &\mathbb{E} \big[ \big( \Delta \hat{\theta}_{PS,3}^{n}(t) \big)^{2} \big] = \mathbb{E} \Big[ \Big( \sum_{m = 1}^{M} \sum_{c = 1}^{C} \sum_{i = 1}^{I} \sum_{k = 1}^{K} \frac{1}{P_{t} MCK \sigma_{h}^{2} \Bar{\beta}_{c}} \Big. \Big. \nonumber \\
    & \hspace{1cm} \Big. \Big. \times \operatorname{Re} \big\{ \big( h_{m,c,k}^{i,n}(t) \big)^{*} z_{c,k}^{i,n}(t) \big\} \Big)^{2} \Big] \nonumber \\
    & = \mathbb{E} \Big[ \sum_{m = 1}^{M} \sum_{c = 1}^{C} \sum_{i = 1}^{I} \sum_{k = 1}^{K} \frac{1}{P_{t}^{2} M^2 C^2 K^2 \sigma_{h}^{4} \Bar{\beta}_{c}^{2}} \Big. \nonumber \\
    & \Big. \hspace{1cm} \times \big( \operatorname{Re} \big\{ \big( h_{m,c,k}^{i,n}(t) \big)^{*} z_{c,k}^{i,n}(t) \big\} \big)^{2} \Big] \nonumber \\
    & = \frac{\sigma_{z}^{2} I}{2 P_{t}^{2} M^2 C^2 K \sigma_{h}^{2}} \sum_{m = 1}^{M} \sum_{c = 1}^{C} \frac{\beta_{m,c}}{\Bar{\beta}_{c}^{2}}.
\end{align}
The same result holds for $N+1 \leq n \leq 2N$. Combining the two results concludes the lemma.
\end{IEEEproof}
Combining the results in Lemmas \ref{convproof11}, \ref{convproof12}, and \ref{convproof13} and applying Assumption \ref{assumption2} with (\ref{eq: useriteration}) completes the proof of Lemma \ref{convproof1}.

\bibliographystyle{IEEEtran}  
\bibliography{references}  

\end{document}